\documentclass[acmtog,nonacm]{acmart}

\setcopyright{none}
\acmDOI{}
\acmISBN{}
\acmConference{}{}{}{}
\acmBooktitle{}

\AtBeginDocument{%
  }

\usepackage{graphicx}
\usepackage{amsmath}

\usepackage{amssymb}
\usepackage{booktabs}
\usepackage{xcolor}
\usepackage{natbib}
\usepackage{listings}
\usepackage{hyperref}
\usepackage{cleveref}
\crefname{section}{Sec.}{Secs.}
\Crefname{section}{Section}{Sections}
\Crefname{table}{Table}{Tables}
\crefname{table}{Tab.}{Tabs.}
\crefname{figure}{Fig.}{Figs.}
\usepackage{stackengine}
\usepackage{tikz}
\usepackage{array}
\usepackage{subcaption}
\usepackage{svg}
\usepackage[dvipsnames]{xcolor}
\usepackage{xspace}
\usepackage{caption}

\usepackage[table,xcdraw]{xcolor}
\usepackage[normalem]{ulem}
\useunder{\uline}{\ul}{}

\usepackage{enumitem}

\lstdefinelanguage{json}{
    basicstyle=\ttfamily\small,
    numbers=left,
    numberstyle=\tiny,
    stepnumber=1,
    numbersep=5pt,
    showstringspaces=false,
    breaklines=true,
    frame=single,
    backgroundcolor=\color{gray!10},
}

\newcommand{\jsoncolor}[2]{\textcolor{#1}{#2}}

\newcommand{\LatentNoisyChannels}{z_t}
\newcommand{\LatentMaskChannels}{z_{mask}}
\newcommand{\LatentCleanChannels}{z_c}

\newcommand{\MemoryLatent}{z^M_c}

\newcommand{\DenseLatentState}{S}

\newcommand{\LatentChannels}{C}
\newcommand{\LatentHeight}{H}
\newcommand{\LatentWidth}{W}
\newcommand{\MemorySize}{M}

\newcommand{\EntityBank}{\mathcal{E}}

\newcommand{\MinMatchThreshold}{\tau_{minmatch}}
\newcommand{\RedundancyThreshold}{\tau_{redundant}}

\newcommand{\KeepScore}{\ensuremath{s^{\text{keep}}}}
\newcommand{\MaxDinoSimilarityScore}{\ensuremath{s^{\text{DINO}}_{\max}}}
\newcommand{\ClipSimilarityScore}{\ensuremath{s^{\text{CLIP}}}}
\newcommand{\NumberOfKeptTokens}{\ensuremath{N_{\text{keep}}}}
\newcommand{\CandidateEntry}{\ensuremath{c^{e}_i}}

\newcommand{\MethodName}{\emph{EM-Vid}\xspace}

\newcommand{\eg}{e.g.\xspace}

\usepackage{xr}
\begin{document}


\title{\textbf{EM-Vid}: Training-Free \textbf{E}ntity-Centric \textbf{M}emory for Efficient and Consistent Multi-Shot \textbf{Vid}eo Generation}


\author{Jente Vandersanden}
\affiliation{%
  \institution{Max Planck Institute for Informatics}
  \city{Saarbrücken}
  \country{Germany}
  \institution{and Adobe Research}
  \city{London}
  \country{UK}
  }
\email{jvanders@mpi-inf.mpg.de}

\author{Matheus Gadelha}
\affiliation{%
  \institution{Adobe Research}
  \city{San Jose}
  \country{USA}
}

\author{Chun-Hao P. Huang}
\affiliation{%
  \institution{Adobe Research}
  \city{London}
  \country{UK}
}

\author{Hyeonho Jeong}
\affiliation{%
  \institution{Adobe Research}
  \city{London}
  \country{UK}
}

\author{Yulia Gryaditskaya}
\affiliation{%
  \institution{Adobe Research}
  \city{London}
  \country{UK}
}

\renewcommand{\shortauthors}{Vandersanden et al.}

\begin{abstract}

Multi-shot video generation requires maintaining a consistent appearance of recurring entities across shots while remaining faithful to shot-specific text prompts. Recent autoregressive methods reuse previously generated frames as memory. However, full-frame storage entangles persistent entity information with transient scene context, leading to irrelevant information leakage and high computational cost. We propose an entity-centric memory in the form of an entity-indexed bank of latent patches. We introduce sparse token conditioning compatible with pretrained models, restricting self-attention to entity-relevant tokens and reducing computational cost. To support this, we introduce a structured multi-shot script format. We additionally propose a budgeted memory update strategy to maintain a compact, evolving memory. Finally, we equip the entity representation with a noise-injection mechanism that enables fine-grained appearance control, preventing leakage of irrelevant information. Our method improves prompt adherence and efficiency while preserving subject consistency.
\end{abstract}




\begin{teaserfigure}
    \centering
    \includegraphics[width=\textwidth]{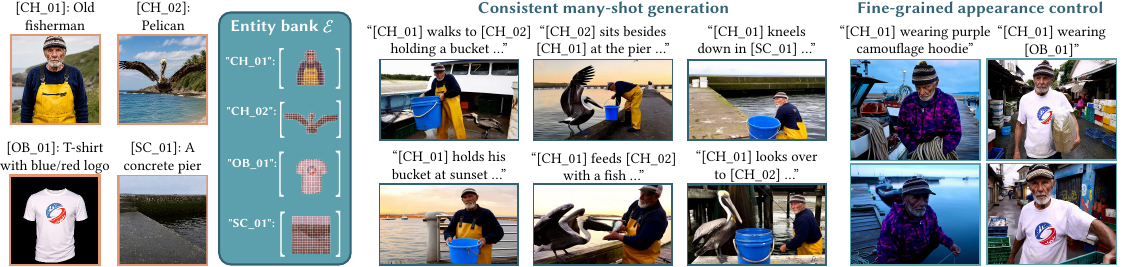}
    \captionsetup{skip = 2pt}
    \caption{
        We introduce a token-efficient entity-centric memory representation, the \emph{entity bank}, which separates persistent entity information from transient scene context. 
        The entity bank supports both user-provided references and updates from generated content, storing the appearance of characters, objects, and scenes (left). 
        Conditioned on this representation, a video generation model produces multi-shot videos in which entities such as the fisherman, pelican, and props remain consistent across diverse prompts and scenes (middle). 
        Our method also enables fine-grained appearance control, allowing specific attributes (e.g., the fisherman’s clothing) to be effectively modified in a new shot through text or additional visual references (right).
        To enable efficient memory querying, we rely on a \emph{structured script} in which persistent subjects and scenes are assigned unique identifiers and explicitly referenced in each shot prompt, \eg [CH\_01].
    }
    \label{fig:teaser}
\end{teaserfigure}

\maketitle

\section{Introduction}
\label{sec:introduction}

Generating multi-shot videos is essential for coherent, long-form visual narratives.
In this setting, several key requirements must be satisfied: visual consistency across shots to preserve narrative continuity, faithful adherence to textual prompts to maintain alignment with user intent, the ability to incorporate newly generated information, and computational efficiency.
Despite recent progress, jointly satisfying these objectives remains challenging. 
To address this, we leverage large pre-trained models and introduce an inference-time modification that improves efficiency and adherence to text prompts while reducing interference from irrelevant information.

\begin{figure} 
\setlength{\tabcolsep}{.1cm} \renewcommand{\arraystretch}{0} \input{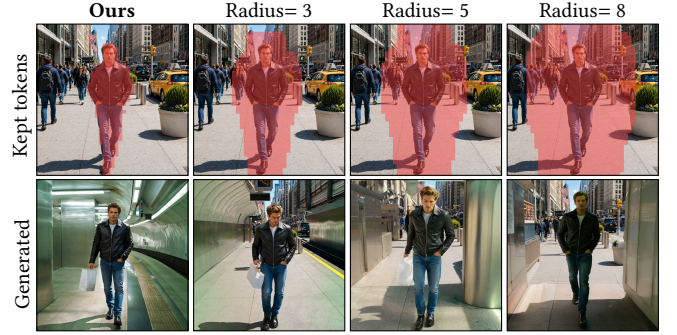} 
\captionsetup{skip = -2pt} 
\caption{Generated shots with varying context sizes for the prompt \textit{``A man walking along an empty subway platform with metallic walls\ldots''}. The top row shows the conditioning frames, with tokens included in the conditioning highlighted in red. As the number of background tokens increases, additional background artifacts (e.g., traffic lights, cars) emerge in the generated video, despite not being specified in the prompt. This demonstrates that incorporating irrelevant information into the conditioning signal can lead to unintended leakage of such elements into the generated output. 
} 
\label{fig:background_leaking} 
\vspace{-0.3cm}
\end{figure}

Early approaches generate all shots jointly \cite{guo2025long}, enabling global interactions that achieve good consistency and adherence to text prompts; however, they are computationally expensive and difficult to scale. 
To improve efficiency, autoregressive methods \cite{huang2025self,zhang2025packing, wu2025pack, an2025onestory, zhang2025pretraining, jiang2025lovic} generate shots sequentially within a limited temporal window, leveraging recent context while reducing computational cost. Nevertheless, their restricted horizon often leads to forgetting of earlier content and accumulated errors over time, degrading long-range consistency. 
A more recent approach \cite{zhang2025storymem} selects representative frames from previously generated shots (memory frames) to guide generation, offering a trade-off between visual coherence and computational efficiency.
While effective in modeling visual history, it is limited by its use of holistic frames as memory. 
This entangles persistent entity appearance with transient scene context, which can lead to irrelevant artifacts leaking into the generated video (see \cref{fig:background_leaking}) and incurring higher-than-necessary computational cost due to the quadratic complexity of self-attention over tokens.
Moreover, we observe that semantically similar prompts can cause the model to favor memory over text, `copy-pasting' the stored appearance instead of faithfully following the prompt.
Motivated by these observations, we propose \textbf{\MethodName}, an entity-centric memory representation and training-free inference mechanism for memory-frame-guided video generation. 
Instead of storing memory as holistic frames, \MethodName represents memory as an entity-indexed bank of sparse latent patches associated with individual entities, such as subjects or scene environments (\cref{fig:teaser}).
Importantly, we observe that although removing irrelevant tokens helps reduce information leakage, some tokens inevitably correspond to boundary patches that still retain irrelevant visual content.
To achieve finer-grained appearance control, we inject pixel-space noise to the irrelevant regions of these boundary patches. 

To make entity-specific retrieval tractable at inference time, we adopt a structured story script format in which persistent subjects and scenes are assigned unique identifiers and explicitly referenced in each shot prompt (\cref{fig:teaser}).
This representation makes the set of shot-relevant entities directly available during generation. \MethodName{} can therefore condition the model only on the memory entries associated with those entities or their parts, reducing both computational cost and interference from irrelevant visual information.

Incorporating such sparse memory into a pretrained memory-frame-guided video generation model is non-trivial, since the original architecture expects dense full-frame latent representations. 
To preserve compatibility with the pretrained model's computation scheme and memory layout, we scatter the selected sparse latent patches back into their original spatial locations within the full latent grid before applying convolution-based patchification. 
The resulting tokens are then sparsified again so that self-attention operates only on entity-relevant content. 
This design preserves the original video generation model computation pathway while substantially reducing unnecessary attention over irrelevant regions.

To support strong consistency, we introduce an entity memory update mechanism that integrates newly observed entity information over time while enforcing a fixed memory-token budget for long multi-shot generation. 

We summarize our contributions as follows: 
\begin{enumerate}[topsep=2pt]
    \item We introduce \MethodName, an entity-indexed memory representation for memory-frame-guided generation that disentangles persistent entity information from transient context by storing sparse entity-associated latent patches. This improves computational efficiency and also enhances prompt adherence, particularly in challenging cases where semantic changes are subtle relative to previous generations.
    \item We propose a training-free sparse-token conditioning mechanism that enables a pretrained memory-frame-guided transformer model to operate on shot-relevant memory tokens only, 
    while maintaining compatibility with the model's original memory layout and computation scheme.
    \item  
    We propose a progressive entity-memory update mechanism that detects entities in newly generated shots, extracts their latent patch entries, and maintains a compact entity bank through relevance- and budget-aware pruning. 
    \item We introduce a noising-based strategy that suppresses irrelevant boundary information when token pruning is too coarse, weakening reference cues that would otherwise preserve the original appearance.

    \item We evaluate \MethodName on multi-shot generation and show improved prompt adherence, computational efficiency, and subject consistency. 
    We also introduce two metrics for cross-shot subject and background consistency that are not biased toward naive appearance copy-paste behavior. 
\end{enumerate}

\section{Related work}
\label{sec:related_work}
Multi-shot video generation requires producing multiple shots with consistent content. 
A central question is how visual information should be represented and propagated across shots.
Existing approaches either generate all shots jointly \citep{meng2025holocine, wang2025multishotmaster, seawead2025seaweed, gao2025seedance, kara2025shotadapter, jia2025moga}, decompose the task into keyframe generation followed by image-to-video synthesis \citep{hu2024storyagent, long2024videostudio,
zhuang2024vlogger, zhao2024moviedreamer, wu2025automated, xiao2025captain}, or generate shots autoregressively while compressing previous shots into a compact history representation \citep{zhang2025packing, jiang2025lovic, an2025onestory}.
Memory-based methods \citep{zhang2025storymem} retain visual evidence from earlier shots and condition future generation directly on that memory. 
Our method follows the last approach, but represents memory as sparse subject-centric visual tokens rather than full keyframes.

\paragraph{Joint generation of all shots}
A direct approach to achieve cross-shot consistency is to generate all shots jointly within a single denoising process. 
These methods typically use per-shot prompts, sometimes together with a global prompt for narrative alignment \citep{meng2025holocine, wang2025multishotmaster}, and enable information sharing across shots via joint self-attention.
They encode shot separation by modifying positional embeddings (e.g., RoPE) to indicate shot boundaries and route textual tokens to their corresponding shot segments, or to position reference subjects within specific shots \citep{seawead2025seaweed,gao2025seedance,wang2025multishotmaster}. 
Attention masking further controls token interactions by restricting attention within shots or selectively enabling cross-shot communication, reducing the cost of full self-attention \citep{kara2025shotadapter,meng2025holocine,wang2025multishotmaster,jia2025moga}.
Our method instead generates shots sequentially using an explicit visual memory, avoiding joint processing of all shots and reducing computational cost.

\paragraph{Keyframe-based approaches}
Multi-shot generation involves first generating a consistent set of keyframes per shot, which are then animated using an image-to-video (I2V) model.
Consistent initialization can be obtained using personalization-based diffusion models \citep{ye2023ip, li2023blip, wang2024ms} or specialized multi-image subject-consistent methods \citep{zhou2024storydiffusion}, followed by I2V models for video synthesis \citep{hu2024storyagent, long2024videostudio, zhuang2024vlogger, zhao2024moviedreamer, wu2025automated, xiao2025captain}.
This modular design leverages pretrained image and video models, but has key limitations: it requires separate models, typically supports only a small number of consistent subjects, and provides limited cross-shot interaction during video generation. 
As a result, newly generated content is usually not fed back into later conditioning, leading to drift and accumulated errors over long narratives.

\paragraph{Autoregressive next-shot generation}
Autoregressive next-shot generation typically maintains a bounded conditioning window through context compression or memory.
Many methods compress prior visual history using efficient context packing or token reduction \citep{zhang2025packing, jiang2025lovic, zhang2025pretraining, an2025onestory}, improving scalability but at the cost of losing or degrading long-range information.
In contrast, memory-frame-based methods retain an explicit bank of frames from previous shots instead of compressing it \citep{zhang2025storymem}.
Our method follows this memory-frame-based formulation but differs in granularity: rather than storing dense frame-level memory, we maintain a sparse entity-level memory that preserves only the most relevant visual information while reducing redundancy and cost.

\paragraph{Subject-to-video generation}
A related line of work aligns generated videos with text prompts and visual subject references.
Methods differ primarily in how the visual conditioning is injected: 
\citet{liu2025phantom} use a dual-branch MMDiT with branch-aligned encoders (CLIP for text, 2D VAE for vision) and support both single- and multi-subject references within a single video.
\citet{chen2025multi} introduce a dedicated personalization cross-attention layer and explicitly bind each reference image to its corresponding entity word in the prompt, enabling open-set multi-subject and background personalization.
\citet{mai2025contextanyone} use an Emphasize-Attention module and Gap-RoPE to inject and isolate reference identity,
and \citet{ma2024magic} specialize in identity-specific human video customization through dedicated identity embeddings. 
These methods share our goal of preserving subject identity under text guidance but operate in a single-shot generation setup: each call produces one video conditioned on fixed reference images, with no mechanism to integrate visual content generated in earlier shots. 
Our method is complementary in this sense: rather than redesigning the conditioning pipeline of a base model, we introduce an entity-centric memory that keeps the context to subject-relevant patches, can be updated as new shots are generated, and remains compatible with pretrained M2V backbones.
Very recently, \citet{chen2025phantom} also addressed the problem of 
information entanglement and leakage of irrelevant content.
However, their method requires fine-tuning on specialized subject-centric data, which is costly and not widely available. In contrast, our approach is training-free.

\section{Method}
Our goal is efficient multi-shot video generation that follows a script while maintaining the consistent appearance of recurring entities, with particular emphasis on reducing memory usage and inference time. 
To this end, we build on prior work that selects frames from previously generated shots, referred to as \emph{memory frames}, to provide context for subsequent shots. 
This approach is motivated by the native image-conditioning capabilities of modern video foundation models \citep{wan2025wan, kong2024hunyuanvideo, hacohen2024ltx}. 
Extending this conditioning mechanism from a single reference image to a set of \emph{memory frames} provides a natural path toward coherent multi-shot generation \citep{zhang2025storymem}.


We introduce inference-time modifications of the memory-to-video model that improve prompt adherence and efficiency in multi-shot generation. 
Namely, instead of conditioning on full keyframes, we construct a structured memory that stores and retrieves visual information at the level of persistent entities---including characters, objects, and scene environments. 
This disentanglement of conditioning information allows each shot to access only the memory relevant to its prompt, reducing redundant context during generation and lowering the computational cost of self-attention while preserving the information required for long-range consistency.

We first describe the underlying model architecture, followed by our memory and script representation and the proposed inference-time modifications.

\begin{figure}
    \centering
    \includegraphics[]{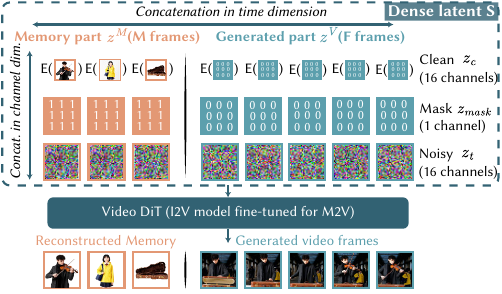}
    \captionsetup{skip = 0pt}
    \caption{
    \textbf{Memory frames to video architecture.} Conditioning is implemented through two complementary mechanisms: time-wise conditioning, where memory frames are encoded and concatenated along the temporal dimension, and channel-wise conditioning, where information is fused via concatenation of noisy latent channels $\LatentNoisyChannels$, binary mask channels $\LatentMaskChannels$, and clean conditioning channels $\LatentCleanChannels$. During diffusion, only $\LatentNoisyChannels$ are iteratively denoised, while $\LatentMaskChannels$ and $\LatentCleanChannels$ remain fixed.
    For more details, see \cref{sec:M2V}.
    }
    \vspace{-0.2cm}
    \label{fig:m2v_overview}
\end{figure}

\subsection{Memory-to-video (M2V) model}
\label{sec:M2V}
Our implementation builds on Wan2.2-I2V \citep{wan2025wan}, using a LoRA fine-tuned variant designed for coherent multi-shot video generation conditioned on memory frames \citep{zhang2025storymem}.
Wan2.2-I2V employs two complementary conditioning mechanisms: \emph{channel-wise conditioning} for keyframe control and \emph{time-wise conditioning} for visual reference propagation (\cref{fig:m2v_overview}).
\emph{Channel-wise conditioning} constructs a unified latent representation by concatenating noisy latent channels $\LatentNoisyChannels$, a binary mask $\LatentMaskChannels$ indicating conditioning versus target frames, and clean conditioning channels $\LatentCleanChannels$.
\emph{Time-wise conditioning} operates along the temporal dimension by concatenating encoded conditioning frames. Each memory image is individually encoded using the 3D VAE of the video model, and the resulting latents are stacked to form $\MemoryLatent \in \mathbb{R}^{\MemorySize \times \LatentChannels \times \LatentHeight \times \LatentWidth}$, which provides visual references across time for generation consistency.

The base I2V formulation supports a single conditioning frame, whereas M2V extends this design to multiple memory frames via the temporal concatenation of $\MemoryLatent$ (\cref{fig:m2v_overview}), enabling multi-reference video generation.
We extend this formulation to a sparser entity-level memory representation. 
To this end, we first need a way to identify which entities each shot requires, which we achieve through our \emph{structured story script} format.

\subsection{Story script format}
\label{sec:script_format}
Our method takes as input a structured story script containing reusable entities and an ordered sequence of shot descriptions.
The format reflects natural story writing: entities are defined once and referenced whenever they reappear.
Entities are grouped into categories such as \texttt{characters}, \texttt{objects}, and \texttt{scenes}, and each is assigned a unique identifier (e.g., \texttt{CH\_01}, \texttt{OB\_01}, \texttt{SC\_01}) together with a short visual description (example is shown in \cref{fig:qual_results} (top)). 

Each shot is described by an \texttt{abstract} and a \texttt{natural} prompt. 
The \texttt{abstract} prompt references entities through their identifiers and is used to retrieve the corresponding memories (example is shown in \cref{fig:qual_results} (top)), while the \texttt{natural} prompt is the fully verbalized text prompt used for generation.  
This allows shot descriptions to act as both generation prompts and implicit memory queries, retrieving only the memory relevant to the current shot while ignoring irrelevant context.
A complete example of the JSON story script format is provided in the supplemental material.
The remaining question is what visual information to store and retrieve for each entity, which we address next.

\begin{figure}
    \centering
       \includegraphics[]{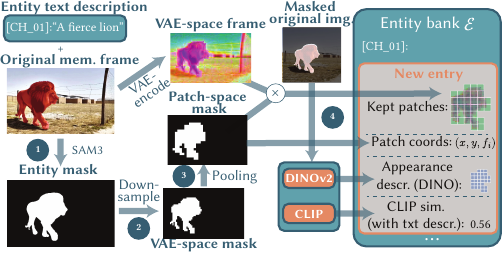}
    \captionsetup{skip = 0pt}
    \caption{
    \textbf{Adding entry to the entity bank.} We introduce an entity bank $\EntityBank$ that stores VAE-space patches for each subject or scene entity. Patches are extracted from memory frames based on overlap with the segmented entity region (pink region in the VAE latent) and stored with their spatial coordinates $(x,y)$ and frame index $f_i$. Each entry is further enriched with region-level descriptors (DINO features and CLIP region-text similarity). New entries are added to update $\EntityBank$ over time as new frames are generated.
    }
    \label{fig:entity_bank_example}
    \vspace{-0.3cm}
\end{figure}

\subsection{An entity-based memory}
\label{sec:subject_memory_representation}


We refine the memory mechanism described in \cref{sec:M2V} by introducing an \emph{entity memory bank} (\cref{fig:entity_bank_example}).
%
This entity-centric representation serves two purposes. By storing only essential entity information, we avoid redundant context during inference that can interfere with self-attention. Second, it provides a compact and efficient representation for maintaining entity consistency across many shots.


\begin{figure*}
\vspace{-0.1cm}
    \centering
    \includegraphics[]{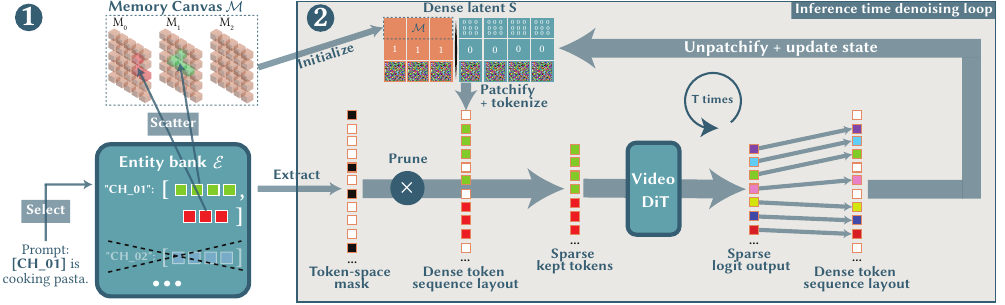}
    \captionsetup{skip = 0pt}
    \caption{
        \textbf{Inference-time sparse memory mechanism.} For each target shot, entity identifiers are extracted from the abstract prompt and used to retrieve the corresponding entries from the entity bank. Sparse latent patches are scattered back to their original spatio-temporal locations to reconstruct a dense latent tensor before patchification (1), preserving consistency with the pretrained video model’s convolution-based tokenization pipeline. After patchification, only tokens corresponding to entity-containing patches are retained and processed by the DiT backbone, reducing self-attention cost and suppressing irrelevant conditioning information. The predicted sparse tokens are then scattered back into the dense token layout (2) before unpatchification, producing the latent-space denoising prediction used to update the full latent state. This process is repeated at every diffusion timestep.
    }
    \label{fig:inference_time_overview}
\end{figure*}

\subsubsection{Entity bank initialization}
\label{sec:subject_bank_initialization}

Assuming a predefined list of subjects and scene descriptions, we initialize an empty entity bank and optionally incorporate user-provided reference images containing characters or scene backgrounds. 

For each reference frame, we start by locating the entities. 
For characters and objects, we use SAM3 \citep{carion2025sam} conditioned on the entity's textual description (\cref{fig:entity_bank_example} (1)); if multiple masks are detected, we take their union. 
For scene entries, we instead detect all foreground elements and define the scene region as their complement, thereby isolating the background scene content. 

Each frame is then encoded into the VAE latent space (\cref{fig:entity_bank_example} (2)), and the corresponding entity masks are downsampled to align with the latent patch grid (\cref{fig:entity_bank_example} (3)), yielding a patch-level indication of which regions belong to the entity. 
In parallel, we extract appearance descriptors of the region by computing DINOv2 embeddings and average-pooling them within the masked area, capturing its visual appearance. 
We further compute CLIP similarity between the masked entity region and its textual description to measure semantic alignment. 
These descriptors are stored alongside the latent patches and later used to guide memory updates (see \cref{sec:update_subject_bank}).

Once all memory frames are VAE-encoded and the corresponding patch-space subject masks are obtained, we construct an entry for the entity bank by storing only the patches that overlap with each entity region. 
For each selected patch, we store its spatial coordinates within the original memory frame, along with the corresponding memory frame index, preserving its position in the global layout.

This yields a compact, subject-centric representation that retains only relevant information from the memory frames. 
Once constructed, the full-frame memory can be discarded, and inference proceeds using only this sparse representation.
\subsubsection{Using the entity bank at inference time}
\label{sec:subject_bank_inference_time}

During inference, for each target shot, we detect entity identifiers (e.g. \texttt{[CH\_01]}) from the \texttt{abstract\_prompt} and retrieve the corresponding entries from the entity bank. 
Our patches reside in the VAE latent space and must be projected to tokens before being passed to the DiT backbone.

In video models, patchification, which partitions the spatio-temporal latent tensor into non-overlapping local regions and maps each region to a feature embedding (a token), is usually implemented using a convolutional operation for efficiency.
Convolutions typically operate on grids, whereas our representation consists of an unstructured subset of patches. 
To preserve consistency with the pretrained model's memory layout and computation scheme, we lift our sparse patches into a regular latent tensor by scattering patches to their original spatio-temporal locations, as shown in \cref{fig:inference_time_overview}~(1). 
Additional discussion on this is provided in the supplemental.

Once projected into tokens, we retain only those corresponding to entity-containing patches. 
This token pruning substantially reduces the computational cost of self-attention and limits the influence of irrelevant information on the generated shots. 

The transformer outputs predictions in token space, which are unpatchified back into VAE latent space to obtain the denoising prediction used by the diffusion process. 
However, since we operate on a sparsified token representation, we first scatter token predictions into the full dense token layout using the binary token mask (\cref{fig:inference_time_overview}~(2)). 
The resulting token sequence is then unpatchified to obtain the corresponding latent-space prediction (e.g., velocity in flow matching), which is used to update the noisy channels of the latent $\DenseLatentState$. 
This procedure constitutes a single denoising step and is repeated across all diffusion timesteps.

\subsubsection{Updating the entity bank}
\label{sec:update_subject_bank}

As new shots are generated, additional information about an entity may become available. 
To maintain cross-shot consistency, we regularly update each entity’s representation in the entity bank. 
Specifically, we extract a small set of candidate keyframes from the newly generated shot based on aesthetic score, following \citet{zhang2025storymem}. 
From these keyframes, candidate entries $\CandidateEntry$ are constructed using the procedure described in \cref{sec:subject_bank_initialization}, where $e$ denotes the entity and $i$ denotes the index of the keyframe from which the entry is derived.
We then compute $\MaxDinoSimilarityScore(\CandidateEntry)$ as the maximum similarity between the candidate’s
appearance descriptor and the appearance descriptors of all entries already stored in the entity bank.
A candidate is accepted if either of the conditions holds: (1) The entity $e$ has no existing entries (i.e., the bank is empty); (2) $\MaxDinoSimilarityScore(\CandidateEntry) \in [\MinMatchThreshold, \RedundancyThreshold]$.
Otherwise, the candidate is rejected, either due to insufficient similarity when $\MaxDinoSimilarityScore(\CandidateEntry) < \MinMatchThreshold$ (to avoid introducing corrupt or unrelated information) or excessive similarity when $\MaxDinoSimilarityScore(\CandidateEntry) > \RedundancyThreshold$ (to avoid redundancy).
When a candidate is accepted, it is appended to the corresponding entity’s entry list.

To prevent unbounded growth of the entity bank, we impose a per-entity token budget. 
While visual redundancy is already controlled via the DINOv2-based filtering above, remaining entries may still exceed the budget. 
In such cases, we select entries greedily based on a relevance-to-cost ratio:
%
    \begin{equation}
      \KeepScore(\CandidateEntry)
      =
      {\ClipSimilarityScore(\CandidateEntry)} / 
           {\NumberOfKeptTokens(\CandidateEntry)} .
  \end{equation}
Here, \ClipSimilarityScore(\CandidateEntry) is the CLIP similarity between the candidate region and the subject description, measuring semantic relevance, while \NumberOfKeptTokens(\CandidateEntry) is the number of memory tokens required by the entry. 
Entries are retained in descending order of $\KeepScore$ until the budget is reached. 
This prioritizes semantically relevant entries while maintaining memory efficiency. To avoid drift from the original appearance, we impose a restriction that the first entry for an entity can never be removed. We provide visualizations of the entity memory evolution in the supplementary material and viewer.




\subsection{Fine-grained entity appearance control via noise injection}
\label{sec:control}
In this section, we introduce a strategy for fine-grained control of entity appearance in generated shots, reducing background leakage and enabling semantically subtle changes in entity attributes.

\subsubsection{Background leakage mitigation}
Once entity-level patches are extracted, those near object boundaries may still contain background information.
This can lead to leakage of the original background into newly generated shots, as illustrated in \cref{fig:bg_noise_ablation}.
To mitigate this, we use image-space masks to identify non-entity regions and inject Gaussian noise into these areas before VAE encoding and storing the patches in our entity bank. 
Gaussian white noise has a flat power spectrum and corrupts all spatial frequencies of a signal, weakening the model’s ability to propagate appearance information from the noised region through attention.



\subsubsection{Selective appearance preservation}
\begin{figure*}
    \centering
    \includegraphics[width=\textwidth]{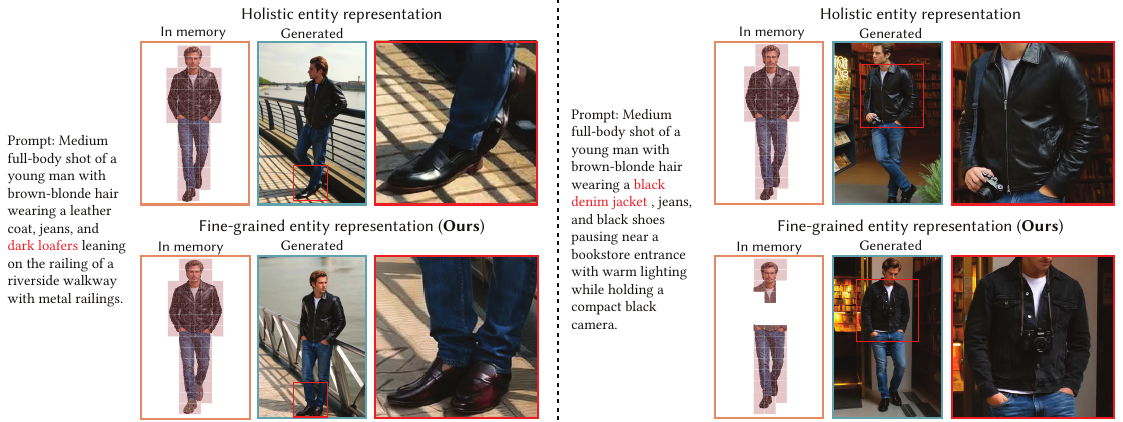}
    \captionsetup{skip = 3pt}
    \caption{\textbf{Fine-grained entity appearance control via noise injection.} 
    We demonstrate the importance of discarding irrelevant tokens and adding noise to boundary tokens when the prompt specifies localized entity modifications, particularly when it is semantically similar to the memory frame. Appearance changes relative to the original entity are highlighted in red in the prompt. With a holistic entity representation, the model may preserve the original visual signal instead of following the prompt, as semantically similar visual tokens can be strongly activated by attention.
    }
    %
    %
    \label{fig:editing_feature_ablation}
\end{figure*}

Additionally, a holistic entity-level representation may not be sufficiently fine-grained when only specific aspects of an entity should be preserved.
For example, attributes such as hairstyle or elements of clothing may need to change in a newly generated shot. 
When the desired attribute differs substantially from that stored in the entity bank, generation is often satisfactory.
However, when the changes are subtle, we observe that the model tends to reproduce the original appearance instead of following the prompt, as shown in \cref{fig:editing_feature_ablation} (top row).
This behavior arises because semantically similar visual memory tokens to the new prompt become strongly activated in self-attention.

To address this, we explicitly segment the regions corresponding to the intended change (e.g., ``shoes'', see \cref{fig:editing_feature_ablation}) and remove patches fully contained within the modification region, retaining only boundary patches.
To prevent undesirable appearance leakage, we apply a similar strategy as for background-containing patches by injecting Gaussian noise into regions whose appearance needs to differ from that stored in the entity bank.

As in the case of background regions, noise injection weakens the model’s ability to propagate outdated irrelevant appearance information through attention.
As a result, generation in this region is primarily guided by the text prompt.

\section{Evaluation}
\label{sec:evaluation}

We evaluate $\MethodName$ along four axes: \emph{visual quality}, \emph{story adherence}, \emph{cross-shot subject and scene consistency}, and \emph{efficiency}. 

\subsection{Test set}
\label{sec:evaluation_test_set}
We perform qualitative and quantitative evaluation on 20 cinematic story scripts, each containing 10 shots, resulting in approximately 50 seconds of generated video per method. 
The scripts include diverse indoor and outdoor environments, recurring entities, varying numbers of subjects per shot, repeated scene revisits, and attribute modifications, enabling evaluation of prompt adherence, cross-shot subject consistency, background-scene consistency, and memory updating. 
The story scripts were generated with GPT-5.4 using prompts for cinematic multi-shot narratives.
Detailed story script statistics are provided in the supplemental material (\cref{tab:data_stats}).

\subsection{Metrics}
To assess \emph{intra-shot visual quality}, we rely on VBench \citep{huang2024vbench} \emph{background consistency}, \emph{image-quality}, and \emph{aesthetic-quality} metrics. 
\emph{Background consistency} is measured by computing CLIP feature similarity across frames to capture the temporal coherence of the background.
\emph{Image quality} is evaluated using MUSIQ \cite{ke2021musiq}, measuring frame-level distortions such as blur, noise, and over-exposure.
\emph{Aesthetic quality} is measured using the LAION aesthetic predictor \cite{laion2022aestheticpredictor}, capturing properties such as layout, color harmony, photorealism, naturalness, and overall artistic quality of video frames.

To evaluate \emph{story adherence}, we use ViCLIP \cite{wang2023internvid} to measure the similarity between each generated shot and its corresponding script.

%


For \emph{Cross-shot Subject Consistency (CSC)}, following \cite{liu2025phantom}, we start by uniformly sampling four frames per generated shot and segmenting the subjects present in them using SAM3. 
%
We then compute CLIP and DINO-based similarity scores. 
For the DINO variant, we first obtain per-frame embeddings by average-pooling subject-masked DINO patch features, while for the CLIP variant, we zero out non-subject regions and encode the masked frame using the CLIP image encoder.
After computing the DINO and CLIP per-frame embeddings over segmented regions, we average them across the frames sampled from each shot and $\ell_2$-normalize the result to obtain a shot-level subject embedding $\mathbf{e}_{s,i}$, where $s$ indexes the subject and $i$ the shot. 
This embedding is used to measure cross-shot subject consistency using cosine similarity.
We refer to this as $CSC_{DINO/CLIP}$.
However, we observe that this formulation favors trivial solutions that "copy-paste" the subject's appearance from reference/memory frames. 
We therefore introduce an additional metric that penalizes the subject consistency score when the silhouettes are overly similar.
%
%
%
Specifically, we also retain the binary subject mask $M_{s,i}$. 
To compare subject shape across shots, we crop each mask to its tight bounding box and resize it to a fixed resolution, placing all masks in a common spatial frame. 
We then compute silhouette similarity as the intersection-over-union (IoU) between the masks:
\begin{equation}
\mathrm{IoU}(M_{s,i}, M_{s,j}) = 
{|M_{s,i}\cap M_{s,j}|}/{|M_{s,i}\cup M_{s,j}|}.
\end{equation}
We apply a penalty only when both visual identity similarity and silhouette overlap are unusually high, which is indicative of copy-paste behavior.
\begin{align}
\mathrm{identity}_{\mathrm{high}} &=
S_{\alpha_1,\alpha_2}
\!\left(
\cos(\mathbf{e}_{s,i}, \mathbf{e}_{s,j})
\right), \\
\mathrm{silhouette}_{\mathrm{same}} &=
S_{\beta_1,\beta_2}
\!\left(
\mathrm{IoU}(M_{s,i}, M_{s,j})
\right),
\end{align}
where $S_{a,b}(\cdot)$ denotes a smoothstep function that smoothly maps values from $0$ to $1$ over the interval $[a,b]$.

We then compute the duplicate risk as
\begin{equation}
\mathrm{duplicate}_{\mathrm{risk}} =
\mathrm{identity}_{\mathrm{high}}
\cdot
\mathrm{silhouette}_{\mathrm{same}}.
\end{equation}

The final cross-shot consistency score is
\begin{equation}
\mathrm{CSC}^{*}_{\mathrm{DINO}/\mathrm{CLIP}} =
\frac{1}{|\mathcal{P}|}
\sum_{(s,i,j)\in \mathcal{P}}
\cos(\mathbf{e}_{s,i}, \mathbf{e}_{s,j})
\left(
1 - \mathrm{duplicate}_{\mathrm{risk}}
\right),
\end{equation}
where $\mathcal{P}$ denotes the set of subject-matched shot pairs. 

Additionally, we introduce a new metric to measure \emph{background scene alignment} across shots, which is not biased towards static background generation. 
We again sample four frames uniformly per shot, segment out foreground subjects and objects, and encode the remaining background regions using either DINOv2 or CLIP. 
A shot-level background embedding $\mathbf{e}_{bg,i}$ is then obtained by averaging the frame-level embeddings, as described above for subjects for the $CSC$ metric.
CLIP captures semantic consistency and is therefore well-suited for comparing scenes with similar textual descriptions but different viewpoints, whereas DINO is more sensitive to fine-grained appearance consistency. 
Together, these metrics provide complementary measures of background consistency.
We compute pairwise similarities between all shot-level background embeddings, producing a set of visual similarity scores. 
We also embed the corresponding scene descriptions from the story script using the CLIP text encoder and compute pairwise similarities between these, producing a set of text-based similarity scores that reflect the expected background similarities (i.e., higher similarity when scene descriptions of the shots are semantically aligned).
The final score is defined as the Spearman rank correlation between the visual background similarity scores and the text-based scene description similarity scores. Higher values indicate better alignment between the generated backgrounds and the underlying scripted scene structure.
We denote these metrics as $BGA_{\mathrm{DINO/CLIP}}$.


Finally, we evaluate \emph{efficiency} in terms of memory token count and inference speed.

\subsection{Results}
\label{sec:results}
\subsubsection{Baselines} 
We compare $\MethodName$ against several baselines, including the vanilla Wan2.2 text-to-video model \citep{wan2025wanopenadvancedlargescale}, 
a full-frame memory baseline, 
\texttt{StoryMem} \cite{zhang2025storymem}, a representative recent autoregressive method for long video generation based on the idea of memory compression with open-source code, 
\texttt{FramePack} \cite{zhang2025packing}, 
and a recent subject-to-video method, \texttt{Phantom} \cite{liu2025phantom}.

We evaluate two versions of our method: one that supports subjects (characters and objects) and scene environments (Ours) and another that only stores memory entries for subjects (Ours w/o Scn).
To enable visual comparison, we generate visual references of subjects that are passed as initial memory to \texttt{StoryMem} and our method. 
These visual references are also passed to \texttt{Phantom}, where each shot is generated independently from the others, as it does not support memory conditioning. 

For FramePack, an autoregressive long-video model that extends an initial frame into a full video, we generate the starting frame with FLUX.2~\cite{flux-2-2025} using
the first-frame prompt of the first shot. We then generate 5-second video segments corresponding to individual shots and update the conditioning prompt at each segment boundary to reflect the prompt of the current shot.

Further implementation details are provided in the supplemental. 

\begin{table}[]
\vspace{-0.1cm}
\captionsetup{skip = 1pt}
\caption{Quantitative comparison of $\MethodName$ against baselines on prompt adherence, background–scene alignment, cross-shot subject consistency, and intra-shot quality metrics (VBench). 
Higher values indicate better performance across all metrics. 
Best results are shown in bold, and second-best are underlined. 
A detailed discussion is provided in \cref{sec:results}.}
\label{tab:quant_results}
\resizebox{\linewidth}{!}{
\begin{tabular}{lcccccc}
\textbf{Metric} &
  \textbf{T2V} &
  \textbf{FrmPck} &
  \textbf{Phntm} &
  \textbf{StrMm} &
  \textbf{Ours w/o Scn} &
  \textbf{Ours} \\ \hline
\multicolumn{7}{c}{\textit{Intra-shot metrics (VBench)}} \\
BG cons. &
  \cellcolor[HTML]{AADDC4}{\ul 0.939} &
  \cellcolor[HTML]{57BB8A}\textbf{0.961} &
  \cellcolor[HTML]{BEE5D2}0.933 &
  \cellcolor[HTML]{FFFFFF}0.915 &
  \cellcolor[HTML]{F5FBF8}0.918 &
  \cellcolor[HTML]{FBFEFC}0.916 \\
Aesthetic &
  \cellcolor[HTML]{57BB8A}\textbf{0.629} &
  \cellcolor[HTML]{FFFFFF}0.511 &
  \cellcolor[HTML]{78C8A1}{\ul 0.606} &
  \cellcolor[HTML]{A4DAC0}0.575 &
  \cellcolor[HTML]{9BD7BA}0.581 &
  \cellcolor[HTML]{9DD8BB}0.580 \\
Imaging &
  \cellcolor[HTML]{57BB8A}\textbf{0.713} &
  \cellcolor[HTML]{FFFFFF}0.594 &
  \cellcolor[HTML]{73C79D}{\ul 0.694} &
  \cellcolor[HTML]{A9DDC4}0.655 &
  \cellcolor[HTML]{81CCA7}0.684 &
  \cellcolor[HTML]{83CDA9}0.682 \\ \hline
\multicolumn{7}{c}{\textit{Prompt adherence}} \\
ViCLIP &
  \cellcolor[HTML]{5CBD8E}0.246 &
  \cellcolor[HTML]{FFFFFF}0.150 &
  \cellcolor[HTML]{ACDEC5}0.199 &
  \cellcolor[HTML]{79C9A2}0.228 &
  \cellcolor[HTML]{5BBD8D}{\ul 0.246} &
  \cellcolor[HTML]{57BB8A}\textbf{0.248} \\ \hline
\multicolumn{7}{c}{\textit{Cross-shot subject consistency}} \\
$CSC_\mathrm{DINO}$ &
  \cellcolor[HTML]{FFFFFF}0.719 &
  \cellcolor[HTML]{FFD666}\textbf{0.936} &
  \cellcolor[HTML]{FFFBF0}0.742 &
  \cellcolor[HTML]{FFF5D8}{\ul 0.776} &
  \cellcolor[HTML]{FFF8E3}0.760 &
  \cellcolor[HTML]{FFFBEF}0.742 \\
$CSC^*_\mathrm{DINO}$ &
  \cellcolor[HTML]{6DC499}0.718 &
  \cellcolor[HTML]{FFFFFF}0.441 &
  \cellcolor[HTML]{67C295}0.729 &
  \cellcolor[HTML]{66C195}0.730 &
  \cellcolor[HTML]{57BB8A}\textbf{0.758} &
  \cellcolor[HTML]{66C194}{\ul 0.731} \\
$CSC_\mathrm{CLIP}$ &
  \cellcolor[HTML]{FFFFFF}0.890 &
  \cellcolor[HTML]{FFD666}\textbf{0.955} &
  \cellcolor[HTML]{FFF7E0}{\ul 0.903} &
  \cellcolor[HTML]{FFFDF5}0.894 &
  \cellcolor[HTML]{FFFDF6}0.894 &
  \cellcolor[HTML]{FFFBEE}0.897 \\
$CSC^*_\mathrm{CLIP}$ &
  \cellcolor[HTML]{5ABD8C}0.875 &
  \cellcolor[HTML]{FFFFFF}0.476 &
  \cellcolor[HTML]{5EBE8F}0.866 &
  \cellcolor[HTML]{66C195}0.846 &
  \cellcolor[HTML]{57BB8A}\textbf{0.881} &
  \cellcolor[HTML]{58BC8B}{\ul 0.879} \\ \hline
\multicolumn{7}{c}{\textit{Alignment between cross-shot generated backgrounds}} \\
$BGA_\mathrm{DINO}$ &
  \cellcolor[HTML]{62C092}{\ul 0.769} &
  \cellcolor[HTML]{FFFFFF}0.600 &
  \cellcolor[HTML]{87CFAC}0.728 &
  \cellcolor[HTML]{85CEAA}0.731 &
  \cellcolor[HTML]{73C79E}0.750 &
  \cellcolor[HTML]{57BB8A}\textbf{0.780} \\
$BGA_\mathrm{CLIP}$ &
  \cellcolor[HTML]{84CEAA}0.682 &
  \cellcolor[HTML]{FFFFFF}0.550 &
  \cellcolor[HTML]{92D3B3}0.667 &
  \cellcolor[HTML]{99D6B8}0.660 &
  \cellcolor[HTML]{80CCA6}{\ul 0.687} &
  \cellcolor[HTML]{57BB8A}\textbf{0.730}
\end{tabular}
\vspace{-0.3cm}
}
\end{table}

\subsubsection{Quantitative evaluation}
In terms of prompt following and \emph{background scene alignment}, $\MethodName$ outperforms all baselines, demonstrating the effectiveness of our strategy in reducing leakage of irrelevant information while preserving consistency.
Our method also consistently improves cross-shot subject consistency, demonstrating stronger preservation of subject identity across shots while still allowing meaningful variation in viewpoint and appearance.
Observe that the naive cross-shot subject consistency metric, CSC, (highlighted with yellow in \cref{tab:quant_results}), favors \emph{FramePack}, which visually exhibits the least variation in subject appearance (\cref{fig:qual_results}).  
Vanilla \emph{Wan2.2-T2V} achieves the highest intra-shot VBench scores, including background consistency, aesthetic quality, and imaging quality. This likely reflects a trade-off between optimizing for long-range story coherence and maximizing per-shot visual fidelity, as reported by \citet{zhang2025storymem}. Nevertheless, our method remains competitive on intra-shot quality metrics while substantially improving story-level consistency and scene alignment.


\subsubsection{Qualitative evaluation}
\begin{figure*}
    \centering
        \includegraphics[width=\textwidth]{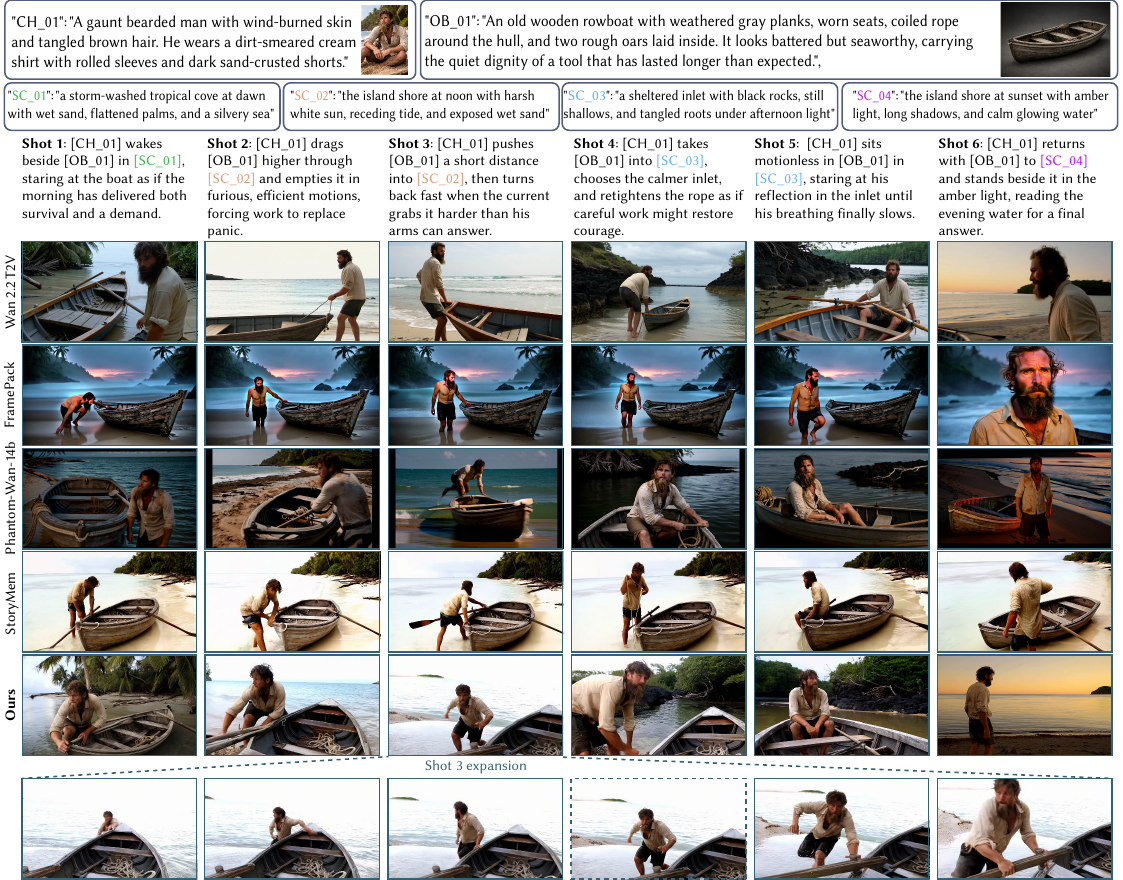}
    \captionsetup{skip = 3pt}
    \caption{
    Qualitative comparison between our method and the baselines \emph{Wan2.2-T2V} \citep{wan2025wanopenadvancedlargescale}, \emph{FramePack} \citep{zhang2025packing}, \emph{Phantom-Wan-14b} \citep{liu2025phantom} and \emph{StoryMem} \citep{zhang2025storymem}. 
    While \emph{Wan2.2-T2V} produces visually appealing individual shots, it struggles to maintain strong subject consistency across shots. 
    \emph{Phantom} is a strong baseline for subject consistency; however, it does not model scene memory. \emph{StoryMem} tends to preserve background appearance across \emph{semantically similar scenes}, whereas our method better balances subject and background consistency with prompt alignment. 
    Full playable versions of these comparisons can be found in the supplementary viewer. 
    }
    \label{fig:qual_results}
\end{figure*}

As shown in \cref{fig:qual_results}, \emph{Wan2.2-T2V} generates high-quality individual shots that closely follow the text prompt; however, without explicit memory of previous generations, it struggles to maintain accurate subject consistency across shots. 
\emph{FramePack}, an autoregressive compressed-memory baseline, often produces unstable shot transitions and struggles to follow individual shot prompts.
%
In the example in \cref{fig:qual_results}, although the script transitions between scenes, the descriptions of \texttt{[SC\_01]}–\texttt{[SC\_04]} remain semantically similar, as they all correspond to visually related shore or beach environments.
As a result, \emph{StoryMem} frequently carries over background content between shots even when the scene should change.
In contrast, our method decouples and disentangles visual memory representations, allowing the model to condition only on the information relevant to the current shot. 
This leads to a better balance between maintaining subject appearance consistency and matching the generated environment to the scene prompt.

\begin{figure*}
    
    \centering
    \includegraphics[width=\textwidth]{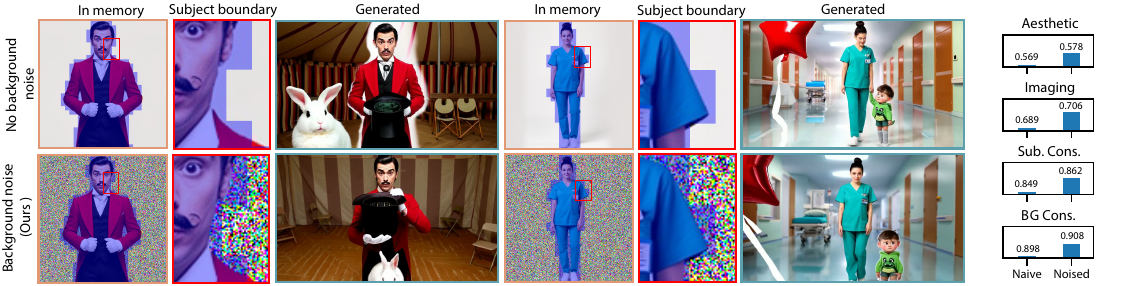}
    \captionsetup{skip = 0pt}
    \caption{
    \textbf{Noise injection reduces background leakage at entity boundaries.}
    Because patch-level selection is coarser than pixel-level segmentation, boundary patches can contain background pixels. These pixels may leak into the generated video, causing
    artifacts such as the white “glow” around the subject. Adding noise suppresses this leakage while preserving the subject, improving visual quality and intra-shot (VBench) metrics (right).}    
    \label{fig:bg_noise_ablation}
\end{figure*}

Next, \cref{fig:bg_noise_ablation,fig:editing_feature_ablation} illustrate how the noise injection strategy described in \cref{sec:control} enables effective appearance propagation in challenging scenarios involving evolving scene backgrounds and semantically similar, but dynamically changing entity attributes. 
As a proof of concept, we only demonstrate changes relative to the initial user-provided references.
However, this approach can be extended to support multi-shot consistency, where multiple attribute-varying versions of the same entity appear across shots.
A detailed discussion is provided in the supplemental.


\vspace{-2pt}
\subsubsection{Efficiency}
\begin{figure*}
    \centering
    
    
    

    \includegraphics[width=\textwidth]{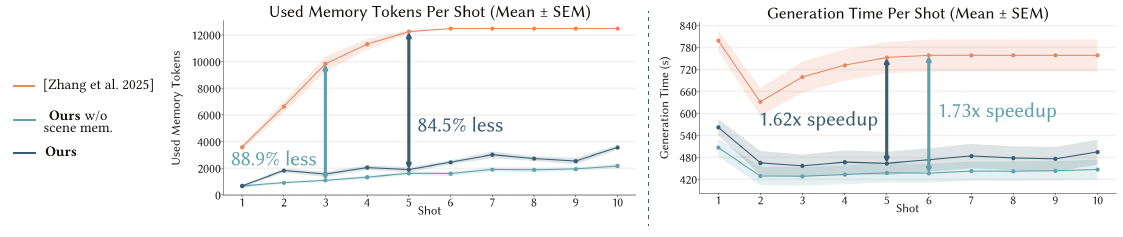}
    \captionsetup{skip = 0pt}
    \caption{
    \textbf{Substantial efficiency gains.}
    Efficiency comparison against the full-frame memory baseline of \citet{zhang2025storymem}. We report memory-token count and per-shot inference time, averaged over the
    20 multi-shot video generation experiments in \cref{sec:evaluation_test_set}. Our sparse memory representation substantially reduces both memory-token usage and
    inference time. Even with scene memory, our method remains efficient by loading only scene entries relevant to the current shot and retaining non-redundant scene
    memories. Interactive visualizations of evolving scene memory are provided in the supplementary viewer.}
    \label{fig:performance_graphs}
\end{figure*}

We show the efficiency of our method against the full-frame memory baseline, StoryMem, in \cref{fig:performance_graphs}. 
On average, we observe up to a $88.9\%$ reduction in memory token usage and up to a $73\%$ inference speedup relative to the dense full-memory baseline.

\section{Conclusion \& Discussion}
\label{sec:conclusion}

\MethodName introduces a memory-efficient, entity-centric approach for multi-shot video generation that improves prompt adherence, subject consistency, and computational efficiency.
The structured entity memory bank, combined with our image-space noising strategy, ensures that only relevant visual evidence is used during generation.
Quantitative and qualitative evaluations demonstrate that \MethodName outperforms existing methods in maintaining subject identity and matching generated environments to scripted prompts, while achieving substantial reductions in memory token usage and inference time. 
Although the approach relies on the SAM3 segmentation model, which may occasionally miss challenging cases, ongoing improvements in segmentation are expected to further enhance robustness. 
Overall, \MethodName provides a scalable and effective solution for consistent, long-form video generation. We will release the source code upon acceptance.


\begin{figure*}
    \centering
    \includegraphics[width=\textwidth]{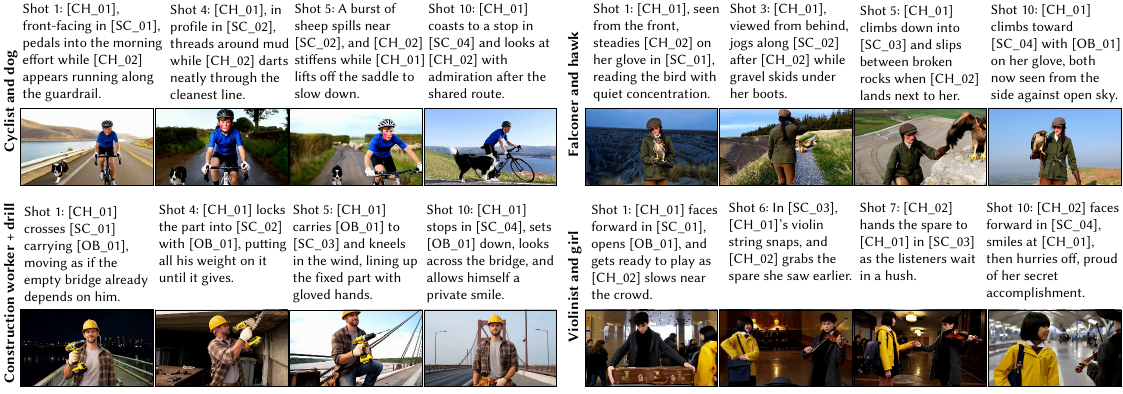}
    \captionsetup{skip = 0pt}
    \caption{
        Additional qualitative results for $\MethodName$ on multi-shot video generation. We observe that $\MethodName$ maintains subject consistency while adhering to the prompt's narrative. More examples and full playable versions can be found in the supplementary viewer.
    }
    \label{fig:additional_results}
\end{figure*}

\clearpage

\bibliographystyle{ACM-Reference-Format}
\bibliography{main}

\clearpage

\appendix

\section{Implementation details}
All our experiments were conducted on a single NVIDIA H100 GPU. In all experiments, we use $\alpha_1=0.88$ and $\alpha_2=0.96$ for the identity similarity thresholds, $\beta_1=0.75$ and $\beta_2=0.90$ for the silhouette IoU thresholds. 
These values were chosen empirically to penalize only pairs exhibiting both unusually high visual similarity and highly overlapping silhouettes, while avoiding unnecessary penalties for naturally consistent subjects across shots.

\section{Example of story script format}
\begin{lstlisting}[language=json]
{
  "story_name": "Boy and Dog",
  "story_overview": "A boy plays with his dog.",
  "characters": [
    {
      "id": §\jsoncolor{red}{"CH\_01"}§ ,
      "short_description": "young boy smiling"
    },
    {
      "id": §\jsoncolor{blue}{"CH\_02"}§,
      "short_description": "small happy dog"
    }
  ],
  "objects": [
    {
      "id": §\jsoncolor{green!60!black}{"OB\_01"}§,
      "short_description": "red ball"
    }
  ],
  "scenes": [
    {
      "id": §\jsoncolor{orange}{"SC\_01"}§,
      "short_description": "green park with pine trees"
    }
  ],
  "shots": [
    {
      "shot_num": 1,
      "abstract_prompt": "§\jsoncolor{red}{[CH\_01]}§ plays with §\jsoncolor{blue}{[CH\_02]}§ in §\jsoncolor{orange}{[SC\_01]}§ using §\jsoncolor{green!60!black}{[OB\_01]}§.",
      "natural_prompt": "young boy smiling plays with small happy dog in green park with pine trees throwing red ball.",
      "first_frame_prompt": "boy throws ball in green park, dog runs."
    }
  ]
}
\end{lstlisting}

\section{Detailed statistics of the used story scripts}
Detailed statistics of the used story scripts are provided in \cref{tab:data_stats}.
\begin{table}[h]
\caption{Detailed statistics of the used story scripts.}
\label{tab:data_stats}
\begin{tabular}{lc}
\hline
\textbf{Statistic}                        & \multicolumn{1}{l}{\textbf{Value}} \\ \hline
\multicolumn{2}{l}{{\ul \textit{Scene Statistics}}}                            \\
Indoor / outdoor shots (\%)               & 25.7 / 74.3                        \\
Avg. scenes per script                    & 4.1                                \\
Avg. shots per scene                      & 2.4                                \\ \hline
\multicolumn{2}{l}{{\ul \textit{Entity Statistics}}}                           \\
Avg. recur. char./obj. per script         & 1.3 (char.) / 1.1 (obj.)           \\
Min / max recur. entities per script      & 2 / 3                              \\
Avg. unique entities per script           & 2.4                                \\
Avg. char. per shot                       & 1.2                                \\
Avg. obj. per shot                        & 0.9                                \\
Max. simultaneous entities in single shot & 3                                  \\ \hline
\multicolumn{2}{l}{{\ul \textit{Temporal Consistency Statistics}}}             \\
Avg. recur. distance between appearances  & 1.1 shots                          \\
Shots containing recur. entities (\%)     & 100                               
\end{tabular}
\end{table}


\section{Entity bank update}
\begin{figure}
    \centering
    \includegraphics[]{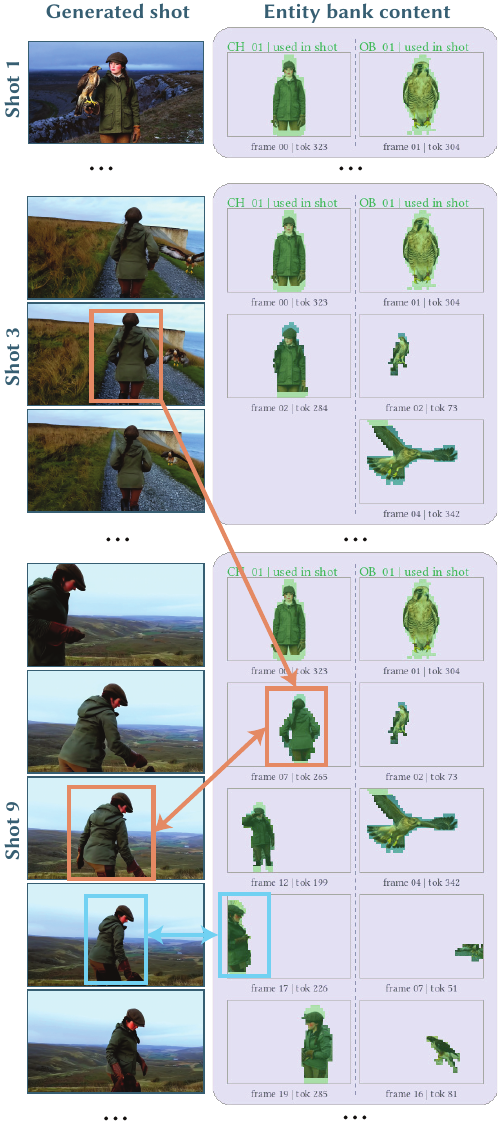}
    \captionsetup{skip = 0pt}
    \caption{
    \textbf{Entity bank evolution.}
  We visualize the evolution of the entity bank throughout generation. The left column shows the generated shots, while the right column shows the corresponding entity bank used for conditioning. The bank can be initialized from user-provided references, as shown here, and is subsequently updated with newly generated observations that provide sufficiently diverse information about an entity. In this example, \texttt{[CH\_01]} is first observed from behind in shot~3, adding a previously unseen view of the character to the bank. When a similar viewpoint reappears in a later shot, the persistent entity representation helps preserve the character’s appearance across shots. 
    }
    \label{fig:entity_bank_update}
\end{figure}

\cref{fig:entity_bank_update} shows how our entity bank is updated over time.


\section{Sparse Entity Patchification}
\label{sec:sparse_vs_dense_patchification}

In designing the sparse inference mechanism, we found that preserving numerical consistency with the pretrained M2V computation scheme and memory layout is important for output quality. 
Our goal is to remove redundant and irrelevant memory content at the DiT level for both efficiency and cleaner conditioning, while keeping the computation as close as possible to the pretrained model’s training-time behavior.
This motivates implementation choices such as scattering sparse subject memory back into a dense layout before patchification and again before unpatchification.

\begin{figure}[t]
    \centering
    \includegraphics[width=0.95\linewidth]{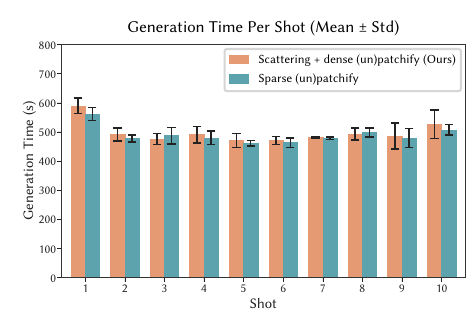}
    \captionsetup{skip = 0pt}
    \caption{ 
    \textbf{Token sparsity, not sparse patchification, drives  speedup.}
    Performance comparison between sparse and dense patchification strategies (see \cref{sec:sparse_vs_dense_patchification}). Sparse patchification is slightly faster in most cases, but the difference between sparse and dense patchification is modest overall. The main performance gain of our method instead comes from conditioning the transformer on only a sparse subset of tokens, which substantially reduces the number of self-attention computations.
    }
    \label{fig:sparse_vs_dense_performance_plot}
\end{figure}

\begin{figure*}[ht]
    \centering
    \includegraphics[]{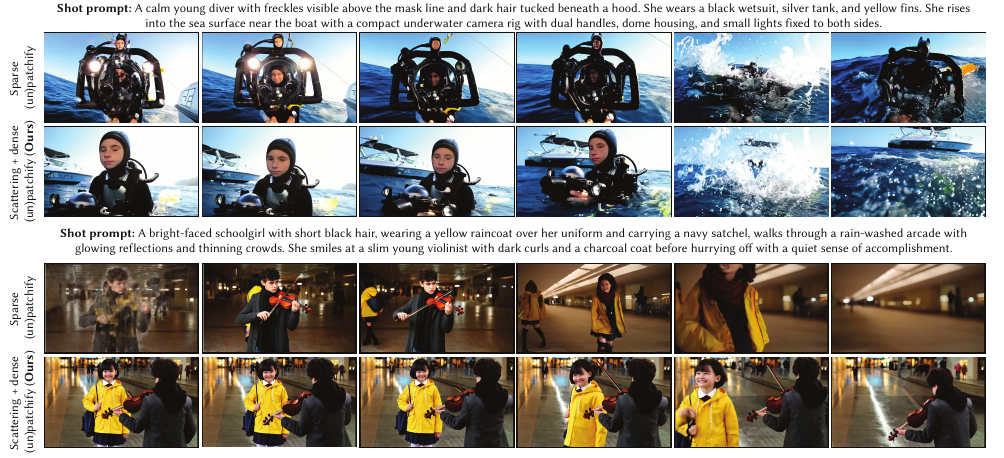}
    \captionsetup{skip = 0pt}
    \caption{
    \textbf{Preserving the pre-trained memory layout matters.}
    Visual comparison between sparse and dense patchification strategies (see \cref{sec:sparse_vs_dense_patchification}). While both strategies produce semantically consistent results, sparse patchification yields noticeable visual artifacts. We attribute this to a train-test mismatch: the pre-trained M2V transformer was trained on patchified latent grids from full memory frames, whereas sparse patchification operates directly on sparse latent patches, altering the ordering of convolution operations. Because these operations are not strictly associative in floating-point arithmetic, this mismatch introduces numerical discrepancies that accumulate over diffusion steps and degrade visual quality. This comparison highlights that our inference design is not arbitrary, but is necessary to preserve the memory layout and numerical behavior expected by the pre-trained M2V model.
    }
    \label{fig:sparse_vs_dense_visual_comparison}
\end{figure*}

To empirically support this design choice, we also implemented a more aggressively sparse variant that directly patchifies and unpatchifies sparse representations using stored patch coordinates, without reconstructing the dense memory layout. 
Although this approach is slightly faster (see \cref{fig:sparse_vs_dense_performance_plot}), it consistently produces lower visual quality (see \cref{fig:sparse_vs_dense_visual_comparison}). 
This suggests that token pruning alone is insufficient to preserve consistency with the computation the pretrained model was optimized for.

We attribute this degradation to a larger train-test gap.
Sparse convolutions necessarily differ from dense 3D convolutions because many intermediate patches that would normally contribute to the computation are omitted. 
Since convolutions involve large sequences of floating-point additions and floating-point arithmetic is not strictly associative (e.g., round(a + round(b+c)) != round(round(a+b) + c), the sparse and dense implementations cannot be numerically identical. 
While the discrepancy introduced by a single convolution is small, these differences accumulate over many layers and diffusion timesteps, eventually affecting generation quality. 
This observation is consistent with prior work emphasizing the importance of minimizing train--test mismatch in diffusion models \citep{huang2025self}.

This consideration motivates our choice of using a pretrained M2V model rather than fine-tuning.
Fine-tuning could compensate for more aggressive modifications to the sparse inference mechanism; however, it would require substantial computational resources as well as large-scale, high-quality, entity-segmented training data tailored to our memory formulation, which is not readily available.
In contrast, M2V models naturally extend I2V models and can be trained or fine-tuned using widely available image data. 
We therefore preserve the pretrained M2V model as-is and modify only the inference-time memory mechanism to selectively condition on relevant entity information. This reduces interference from redundant artifacts and improves efficiency.  

\section{Long video generation}
Our memory representation can also be used for autoregressive long-video generation. Examples are shown in \cref{fig:long_videogen}, with full playable results provided in the supplementary viewer.

\section{Comparison with a keyframe-based approach}
\begin{figure*}[!t]
    \centering
    \includegraphics[]{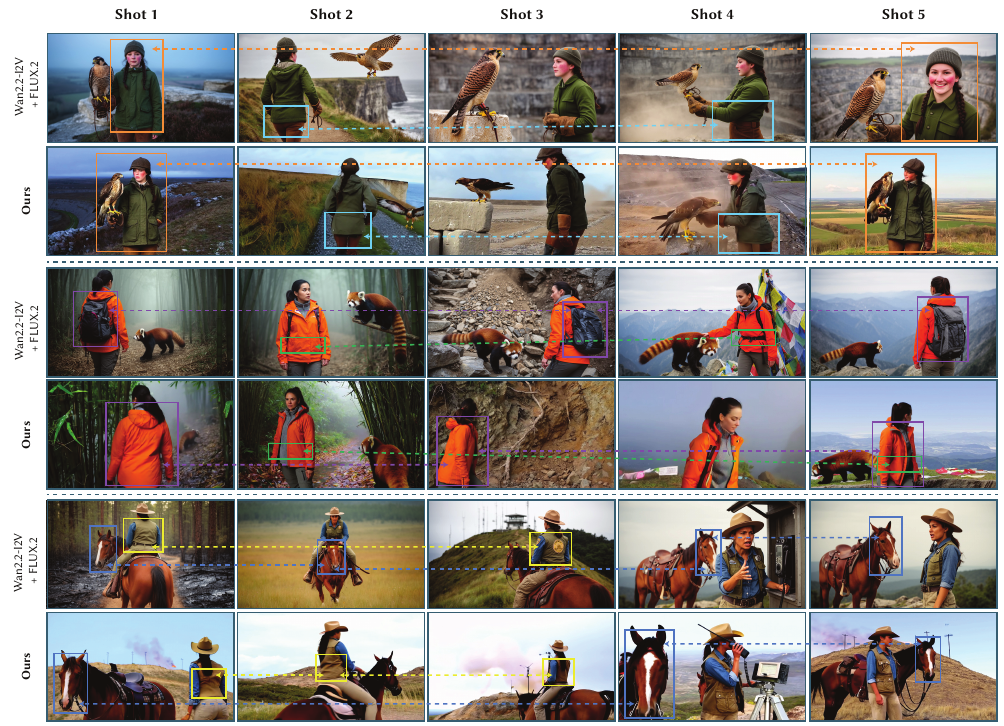}
    \captionsetup{skip = 0pt}
    \caption{\textbf{Comparison to a keyframe-based iterative generation.} 
    We compare our method with a keyframe-based pipeline in which FLUX.2 generates the first frame of each shot, and Wan2.2-I2V animates the resulting keyframes. FLUX.2 is conditioned on the previously generated keyframe and instructed to preserve subject identity. While this approach often maintains local consistency between adjacent keyframes, subject attributes can drift when parts of the subject leave the frame, reappear after several shots, or are observed only from limited viewpoints. The highlighted regions illustrate such long-range inconsistencies in the FLUX.2+Wan2.2-I2V results. In contrast, our method maintains a persistent, updateable entity memory that remains accessible across shots, reducing these inconsistencies.}
    \label{fig:keyframe_based_limitations}
\end{figure*}

A related alternative to memory-augmented video generation is a keyframe-based approach that combines an image generation model with an image-to-video (I2V) model.
We experimented with a representative pipeline in which FLUX.2~\citep{flux-2-2025} iteratively generates the first frame of each shot from a text prompt, conditioned on the previously generated keyframe and instructed to preserve subject identity. 
Each keyframe is then animated using an I2V model.
Although FLUX.2 produces high-quality individual images, this formulation propagates subject information only through local keyframe-to-keyframe conditioning. 
As a result, it lacks an explicit persistent memory of the subjects and remains susceptible to the drift, forgetting, and identity entanglement observed in long-horizon autoregressive video generation (see \cref{fig:keyframe_based_limitations}).






\begin{figure*}[ht]
    \centering
    \includegraphics[width=\textwidth]{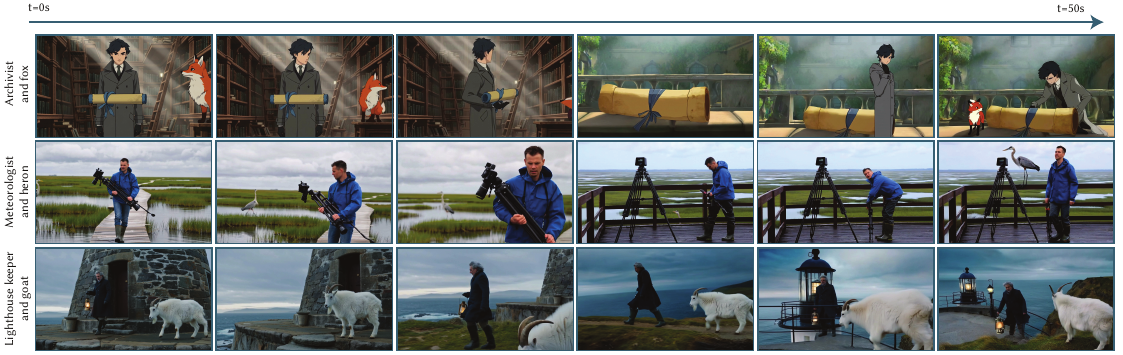}
    \captionsetup{skip = 0pt}
    \caption{\textbf{Long-form video generation.}
    Our memory representation can also be used to autoregressively generate a single long video. We show
    frames from a 50-second video generated by autoregressive rollout using the pre-trained M2V model together with our memory representation. The memory helps preserve subject consistency over long temporal gaps, even when subjects leave the frame and later reappear. For example, in the first row, the fox and archivist exit the fourth frame and return later; in the third row, the lighthouse keeper disappears in the second frame and reappears later in the video. Full playable results are provided in the supplemental viewer.
        }
    \label{fig:long_videogen}
\end{figure*}

\section{Supporting multiple versions of the same entity with evolving attributes}

In the main paper, we demonstrated the importance of token pruning and our noising mechanism. Here, we discuss how this formulation can also naturally be extended to finer-grained appearance control by treating attributes as entities. For example, instead of representing a character \texttt{[CH\_01]} only as a single entity, we can introduce finer-grained entity labels such as
\texttt{[CH\_01\_spiky\_hair]}, \texttt{[CH\_01\_blue\_tshirt]}, \texttt{[CH\_01\_jeans]}, and \newline \texttt{[CH\_01\_white\_sneakers]}. In this case,
the entity bank is organized at a finer granularity, but the underlying mechanisms remain unchanged: each label is treated as an entity, storing sparse memory patches for its corresponding region, and inference selects the entries referenced by the prompt. 
This formulation enables finer-grained control over subject appearance by conditioning on attribute-specific memory entries.

The only additional bookkeeping is to preprocess the script so that prompts refer to these finer-grained labels where appropriate. This allows different attribute versions to be selected across shots while reusing the same budgeted entity-bank update, selection, and sparse inference mechanisms.

\end{document}